# Prediction of the final rank of Players in PUBG with the optimal number of features

Diptakshi Sen[1],Rupam Kumar Roy[2],Ritajit Majumdar[3],Kingshuk Chatterjee[4],Debayan Ganguly[5]

[1] Department of Computer Science and Engineering, University of Calcutta, senn9807@gmail.com
[2] Department of Computer Science and Engineering, University of Calcutta, rupamkrroy98@gmail.com
[3] Advanced Computing and Microelectronics Unit, Indian Statistical Institute, Kolkata,mojutijatir@gmail.com
[4] Department of Computer Science and Engineering, Government College of Engineering and Ceramic Technology,kingshukchaterjee@gmail.com
[5] Department of Computer Science and Engineering, Government College of Engineering and Leather Technology,debayan3737@gmail.com

**Abstract.** PUBG is an online video game that has become very popular among the youths in recent years. Final rank, which indicates the performance of a player, is one of the most important feature for this game. This paper focuses on predicting the final rank of the players based on their skills and abilities. In this paper we have used different machine learning algorithms to predict the final rank of the players on a dataset obtained from kaggle which has 29 features. Using the correlation heatmap,we have varied the number of features used for the model. Out of these models GBR and LGBM have given the best result with the accuracy of 91.63% and 91.26% respectively for 14 features and the accuracy of 90.54% and 90.01% for 8 features. Although the accuracy of the models with 14 features is slightly better than 8 features, the empirical time taken by 8 features is 1.4x lesser than 14 features for LGBM and 1.5x lesser for GBR. Furthermore, reducing the number of features any more significantly hampers the performance of all the ML models. Therefore, we conclude that 8 is the optimal number of features that can be used to predict the final rank of a player in PUBG with high accuracy and low run-time.

## 1 INTRODUCTION

PUBG is an online Battle Royale video game which is a multiplayer shooter game where the players have to fight to remain alive till the end of the game. In this game, a maximum of 100 players are allowed. Players can choose to enter the match solo, duo, or with a team of up to four (squad). Players are dropped empty handed from a plane on one of the four maps at the beginning of the match. Once they land, the players start searching for weapons and armors which are periodically distributed throughout the game. The players then fight one on one and the last player, or team, alive wins the match.

The rank of a player or a team is an important aspect of the game because the rank of a player is the position at which the player or the team gets eliminated and this rank is required to calculate the tier of the player. Machine Learning (ML) based approach to predict the final rank of the players have been studied in some papers [1][2][3]. These papers use different ML based techniques with more than 15 features for this task.

In this paper, we have predicted the rank of the players or teams in PUBG using both previously used algorithms, such as Multiple Linear Regression, LGBM, Random Forest, and some other algorithms as well such as Gradient Boosting Regression(GBR) [4][5], Decision Tree [6], K-Nearest Neighbours (KNN)[7] on a dataset from Kaggle [8]. We find that Light Gradient Boosting. Method (LGBM) [9][10] and GBR[11] provides the highest accuracy of ~91.63% and an MAE of 0.06, which is at par with the earlier studies on this. However, we further show that the number of features can be reduced to 8 (the top 8 features of the correlation heatmap) without hampering the performance of the model. Nevertheless, this reduction in the number of features provides an approximate 1.5x speedup to the empirical runtime of this algorithm. We, further, numerically have shown that reducing the number of features any more have a significant role on the performance of the ML algorithms.
Remaining paper is arranged as follows-
We did *data cleaning in section 2, feature engineering and feature selection in section 3.* Further we have discussed our *findings in result and discussion in section 4* and finally *concluded our paper in section 7* . We have uploaded our code to GitHub, and have provided its link with this paper in *code availability section*.

## 2 Data Cleaning

As discussed earlier, the maximum number of players allowed in SOLO, DUO and SQUAD match types are 1, 2 and 4 respectively. But the dataset has categorized the match types into 16 different types which are variations of these three primary types. We have mapped the number of players into its proper match type (Fig. 1) as follows:

Players in game type j= $\sum_i \textit{players in game type } j_i$ (1)

Where $j_i$ *are the different subformats of the primary game formt j.*

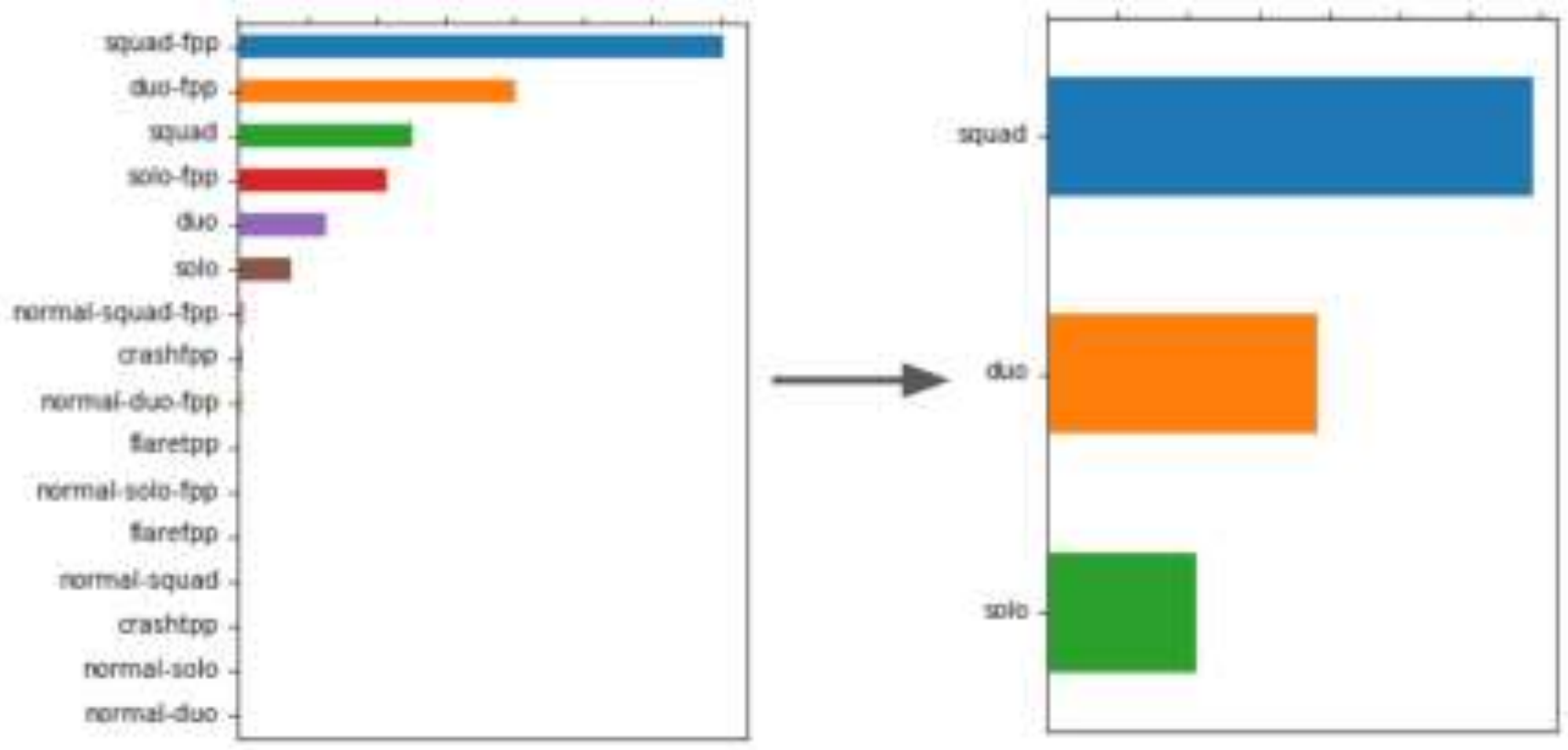

**Fig 1**-Mapping of different match types into core match types.

We have further removed anomalous data from the dataset. The criteria for removal of a data are one or more of the following: (i) Number of players in a team for a particular match type is greater than the allowed number of players, (ii) (possibly offline) players who have either 0 kill and have not covered any distance and have not picked up any weapon.

## 3 Features Engineering and Selection

PUBG allows a maximum of 100 players in a match, but it is not always necessary to have 100 players. If there are 100 players in a match then it might be easier to find and kill enemies as compared to 90 players. Using this notion, we have normalised the features as

$$\text{FEATURES}*\frac{(100-\text{number of players})}{100}+1 \quad (2)$$

This provides a higher score to a player for their achievement when the total number of players is less.

Furthermore, we have created some new compound features by combining existing features. These features lead to a higher accuracy of the ML algorithms. The new features that we have created are

1.Assist_Revive= Assist and revive are the part of teamwork so we have
2.taken both as a single parameter ie., Assist+Revive
3.Total_Distance= Total distance is the distance travelled by the players by walking and swimming i.e., walk distance+swim distance
4.Players_in_a_team= Number of players in team(based on groupID)
5.Headshot/kill= Number of headshot per kill

All these features have been normalized for the prediction purpose.

All the attributes don’ t have equal impact on the final rank of a player. So we have performed feature selection to select those features which will affect the result. In Fig.3 we show the correlation heatmap[12] which is used to select the top 14 features which have high correlation with the target variable.

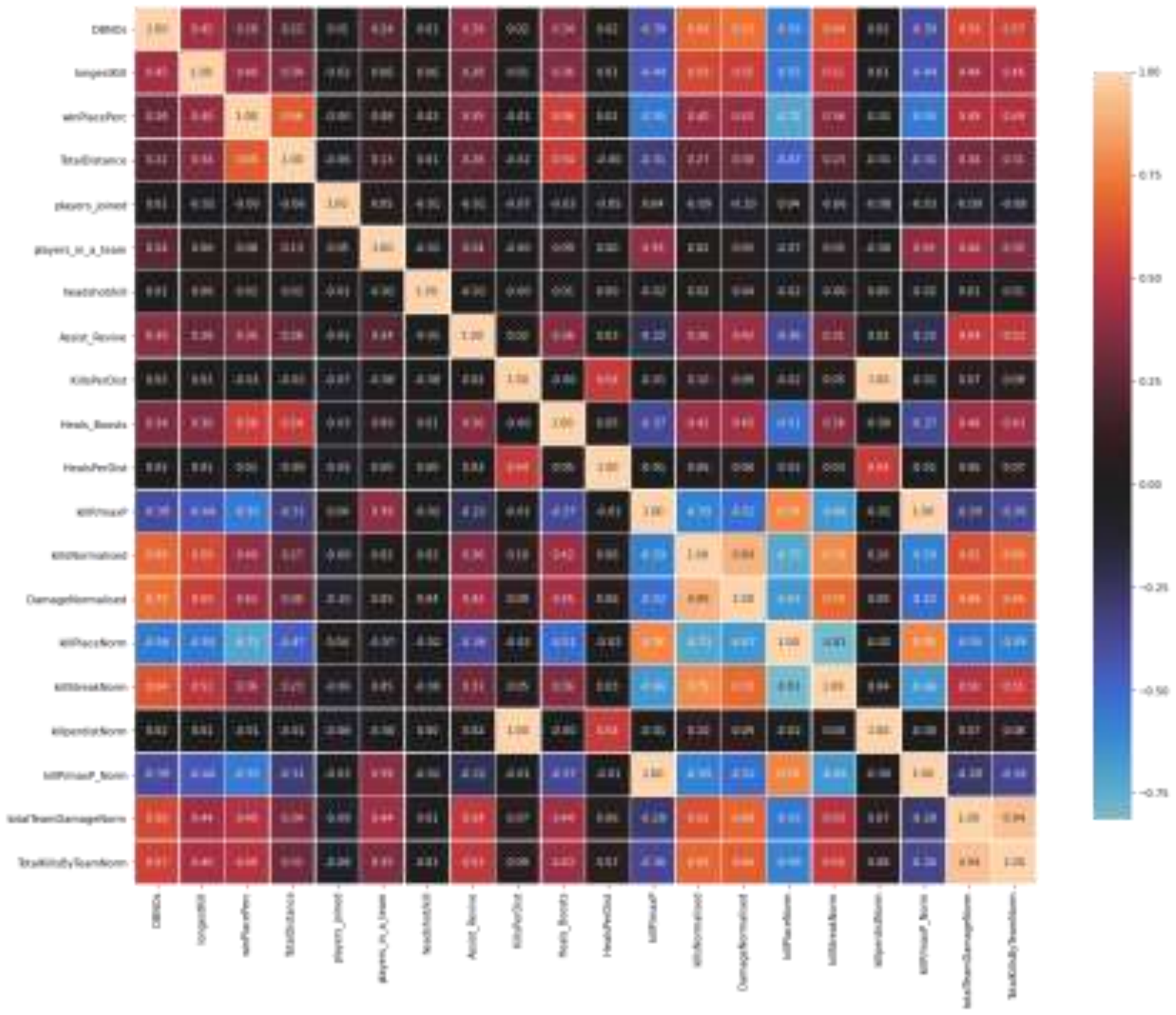

***Fig 2-*** Correlation matrix representing the correlation of the features

We have further studied how reducing the number of features affects the accuracy and MAE of the models. In order to do so, we have varied the number of features from 5 to 8 which are the subsets of previously taken 14 features. We see that upto 8 features, the performance remains more or less steady, but drops significantly for a lower number of features.

## 4 Result and Discussion

We have varied the number of features from 14 to 5 while applying several ML models. We have closely observed the accuracy achieved , empirical time taken and the reduction of MAE while varying the number of features .We have shown our results by comparing the MAE, accuracy and empirical time taken by various methods for a particular set of features. In Table 2-4, we explicitly show the actual features considered and the accuracy, time and MAE for different ML algorithms with 14, 8 and 7 features respectively. Note that all these are standard PUBG features, and we therefore do not explain them further. However, as their names suggest, each of these features have been normalized as discussed in Sec 3. Henceforth, in Fig. 4 and 5 we respectively show the accuracy and the empirical time required for different ML models as the number of features is varied from top 14 to top 5.

14 FEATURES: The used features are:-
DBNOs, killPlaceNorm, killStreakNorm, longestKill, TotalDistance, killperdistNorm, HealsPerDist, Assist_Revive, killP/maxP_Norm, totalTeamDamageNorm, TotalKillsByTeamNorm, killsNormalised, DamageNormalised

***Table 1 (*** Comparison of MAE ,Accuracy ,Time of different models with 14 features)

| Models | *LGBM* | *RANDOM FOREST* | *GBR* | *DECISION TREE* | *KNN* | *RIDGE* | *LASSO* | *LINEAR REGRESSION* |
|---|---|---|---|---|---|---|---|---|
| MAE | 0.063 | 0.065 | 0.062 | 0.088 | 0.100 | 0.110 | 0.115 | 0.110 |
| ACCURACY | 91.26% | 90.45% | 91.63% | 82.59% | 79% | 75% | 73% | 75% |
| TIME(in sec) | 12.73 | 9.20 | 23.37 | 1.38 | 0.79 | 0.03 | 0.30 | 0.05 |

8 FEATURES: The used Features are -
TotalDistance, TotalKillsByTeamNorm, killsNormalised, DamageNormalised, Heals_Boosts, killPlaceNorm, killP/maxP_Norm, longestKill

***Table* 2 *(*** Comparison of MAE ,Accuracy ,Time of different models with 8 features)

| Models | *LGBM* | *RANDOM FOREST* | *GBR* | *DECISION TREE* | *KNN* | *RIDGE* | *LASSO* | *LINEAR REGRESSION* |
|---|---|---|---|---|---|---|---|---|
| MAE | 0.067 | 0.069 | 0.065 | 0.091 | 0.093 | 0.126 | 0.128 | 0.13 |
| ACCURACY | 90.01% | 89.49% | 90.54% | 81.20% | 81% | 69% | 69% | 69% |
| TIME (in sec) | 8.89 | 6.86 | 15.38 | 0.91 | 0.54 | 0.02 | 0.06 | 0.03 |

7 FEATURES: The used features are:-
TotalDistance, TotalKillsByTeamNorm, killsNormalised, DamageNormalised, Heals_Boosts, killP/maxP_Norm, longestKill

***Table 3(*** Comparison of MAE ,Accuracy ,Time of different models with 7 features)

| Models | LGBM | RANDOM FOREST | GBR | DECISION TREE | KNN | RIDGE | LASSO | LINEAR REGRESSION |
|---|---|---|---|---|---|---|---|---|
| MAE | 0.089 | 0.078 | 0.081 | 0.103 | 0.116 | 0.144 | 0.146 | 0.14 |
| ACCURACY | 84.30% | 86.26% | 86.23% | 75.65% | 73% | 63% | 63% | 63% |
| TIME(in sec) | 6.76 | 6.71 | 14.07 | 0.86 | 0.55 | 0.45 | 0.18 | 0.10 |

From the above tables 1-3 we have observed that LGBM and GBR are giving the best result with respect to the MAE and accuracy. From table 1 and table 2 we can say that there is a nominal change in MAE and accuracy for 8 features as that of 14 features although empirical time for the former is much lesser than 14 features. We have further tried to reduce the empirical time by reducing the number of features; but there is a significant degradation of accuracy and the MAE (observed from table 3) and the trend persists (as illustrated in fig 3 and fig 4)

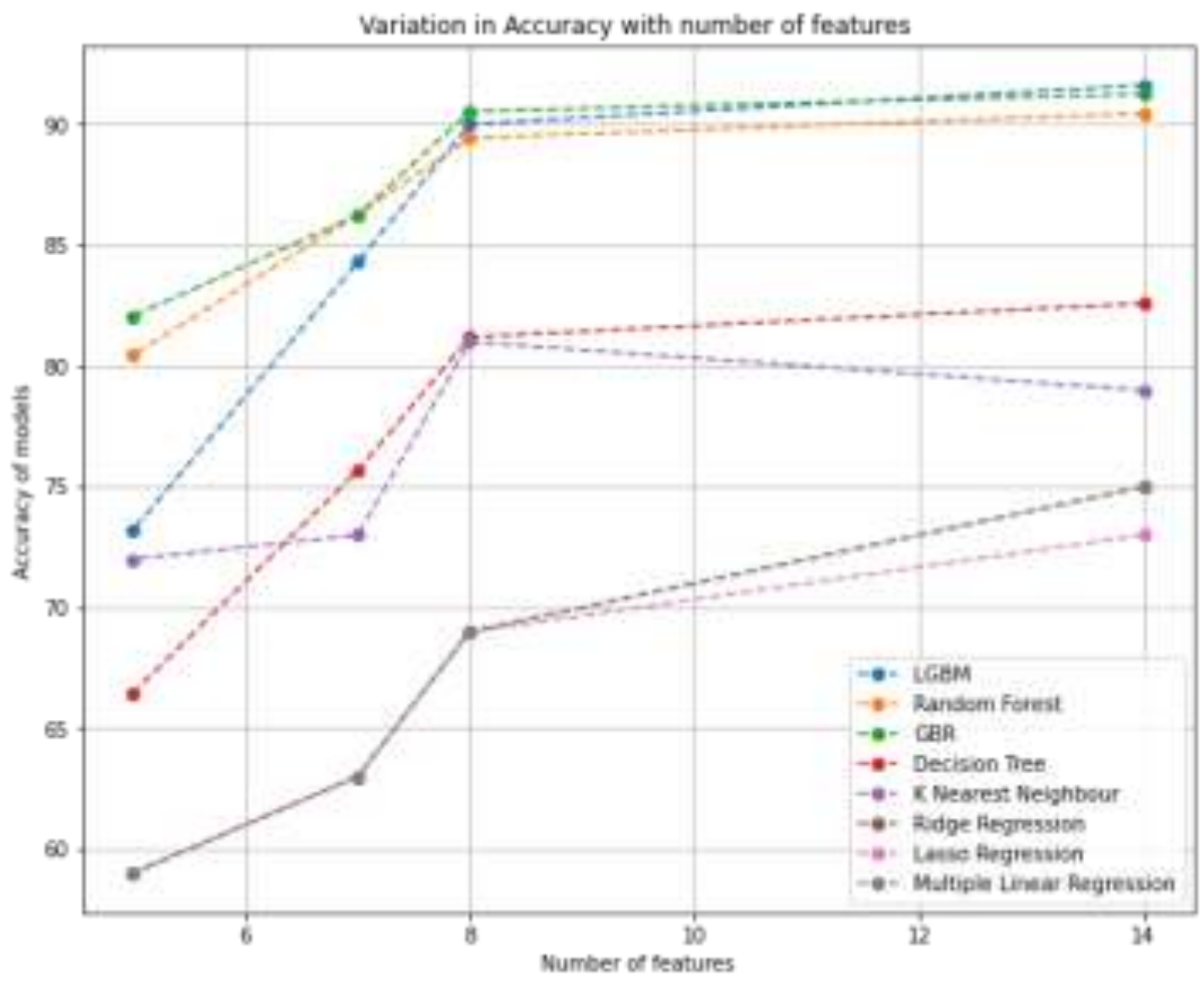

**Fig 3-** Accuracy of the models vs the number of features

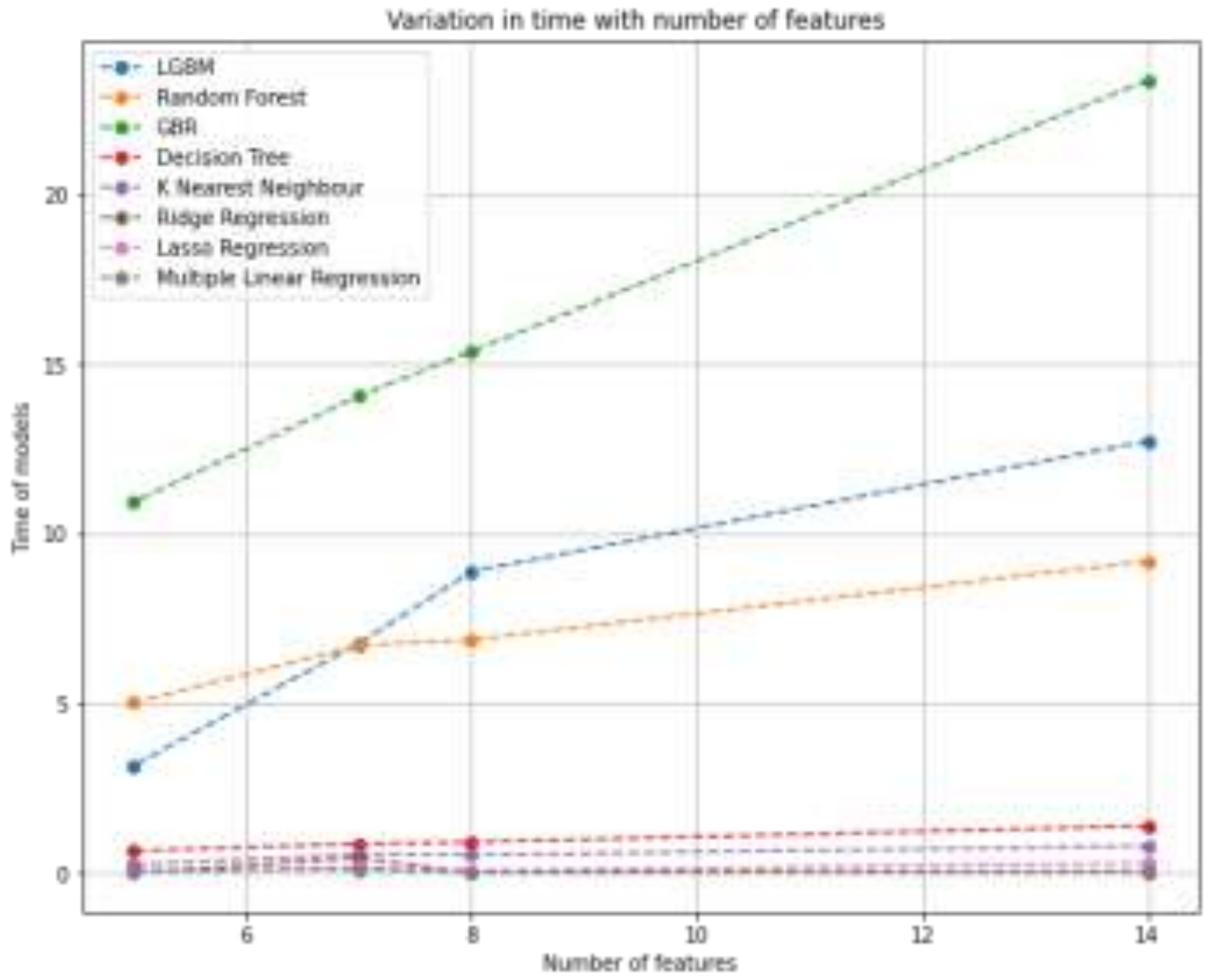

**Fig 4**- Empirical time taken by different features

## 7 Conclusion

In this paper, we have used several ML models to predict the final placement or rank of the players in the PUBG. Out of these models LGBM and GBR have given the best result. It can be concluded that 8 features are more preferable than 14 features because it reduces the empirical runtime of the ML algorithm. Furthermore, 8 is the lower threshold, since further reduction in the number of features significantly degraded the performance. Eventually, although this study is based on PUBG, the technique can be extended to any multiplayer game by using suitable features.

## Code Availability

Code is available at this link-
https://github.com/Diptakshi/PUBG

### References

[1] Brij Rokad, Tushar Karumudi, Omkar Acharya , Akshay Jagtap , " Survival of the Fittest in Player Unknown' s BattleGrounds" , arXiV:1905.06052,2019.
[2] Sampriti Chatterjee, " PUBG Data Analysis using Python," 2020 [online] . Available: https://www.mygreatlearning.com/blog/pubg-data-analysis-using-python/
[3] Wenxin Wei, Xin Lu, Yang Li , " PUBG:A Guide to Free Chicken Dinner" ,Stanford University 2018

[4] Yifei Huang, Yuhua Liu, Chenhui Li, “ GBRTVis : online analysis of gradient boosting regression tree” , journal of visualisation 22,125-140, Springer,2019
[5] Procedia Economics and Finance,volume 39, pages 634-64
, Elsevier,2016
[6] Computational Statistics and Data Analysis, volume 38, Issue 4, Pages 367-378, Elsevier, 2002
[7] Annals of Translational medicine, Vol 4, No 11 , 2016
[8]PUBG finish placement prediction https://www.kaggle.com/c/pubg-finish-placement-prediction
[9] Omar, K.B.A, “ XGBoost and LGBM for PortoSeguro’ s Kaggle challenge: A comparison Preprint Semester Project” , 2018
[10]Guolin Ke., Qi Meng, Thomas Finley,Taifeng Wang, Wei Chen,Weidong Ma, Q,Ye,Tie-Yan Liu, “ LightGBM:A Highly Efficient Gradient Boosting Decision Tree” ,NIPS 2017
[11]Jerome H. Friedman, “ Greedy Function Approximation:A Gradient Boosting Machine” , 1999
[12] Bibor Szabo, “ How to create a seaborn correlation heatmap in python?” ,2020